\documentclass{article}

 \PassOptionsToPackage{numbers, compress}{natbib}
 \usepackage[final]{nips_2016}

\usepackage[utf8]{inputenc} 
\usepackage[T1]{fontenc}    
\usepackage{hyperref}       
\usepackage{url}            
\usepackage{booktabs}       
\usepackage{amsfonts}       
\usepackage{nicefrac}       
\usepackage{microtype}      

\usepackage{mystyle}
\usepackage{mathrsfs}
\usepackage{epstopdf}

\usepackage{amssymb, amsthm, epsfig}

\title{The Robustness of Estimator Composition}

\author{
  Pingfan Tang\\
  School of Computing\\
  University of Utah\\
  Salt Lake City, UT 84112 \\
  \texttt{tang1984@cs.utah.edu} \\
   \And
   Jeff M. Phillips\\
  School of Computing\\
  University of Utah\\
  Salt Lake City, UT 84112 \\
   \texttt{jeffp@cs.utah.edu} \\
}

\begin{document}

\maketitle

\begin{abstract}
We formalize notions of robustness for composite estimators via the notion of a breakdown point.  A composite estimator successively applies two (or more) estimators: on data decomposed into disjoint parts, it applies the first estimator on each part, then the second estimator on the outputs of the first estimator.  And so on, if the composition is of more than two estimators.
Informally, the breakdown point is the minimum fraction of data points which if significantly modified will also significantly modify the output of the estimator, so it is typically desirable to have a large breakdown point.  Our main result shows that, under mild conditions on the individual estimators, the breakdown point of the composite estimator is the product of the breakdown points of the individual estimators.
We also demonstrate several scenarios, ranging from regression to statistical testing, where this analysis is easy to apply, useful in understanding worst case robustness, and sheds powerful insights onto the associated data analysis.
\end{abstract}

\section{Introduction}
Robust statistical estimators~\cite{HRRS86,Huber81} (in particular, resistant estimators), such as the median, are an essential tool in data analysis since they are provably immune to outliers.  Given data with a large fraction of extreme outliers, a robust estimator guarantees the returned value is still within the non-outlier part of the data.
In particular, the roll of these estimators is quickly growing in importance as the scale and automation associated with data collection and data processing becomes more commonplace.  Artisanal data (hand crafted and carefully curated), where potential outliers can be removed, is becoming proportionally less common.  Instead, important decisions are being made blindly based on the output of analysis functions, often without looking at individual data points and their effect on the outcome.  Thus using estimators as part of this pipeline that are not robust are susceptible to erroneous and dangerous decisions as the result of a few extreme and rogue data points.

Although other approaches like regularization and pruning a constant number of obvious outliers are common as well, they do not come with the important guarantees that ensure these unwanted outcomes absolutely cannot occur.

In this paper we initiate the formal study of the robustness of composition of estimators through the notion of breakdown points.  These are especially important with the growth of data analysis pipelines where the final result or prediction is the result of several layers of data processing.  When each layer in this pipeline is modeled as an estimator, then our analysis provides the first general robustness analysis of these processes.

The \emph{breakdown point}~\cite{Ham71,DG07} is a basic measure of robustness of an estimator.  Intuitively, it describes how many outliers can be in the data without the estimator becoming unreliable.  However, the literature is full of slightly inconsistent and informal definitions of this concept.
For example:
\begin{itemize}
\item Aloupis~\cite{Alo06} write
``the breakdown point is the proportion of data which must be moved to infinity so that the estimator will do the same.''

\item Huber and Ronchetti~\cite{HR2009} write
``the breakdown point is the smallest fraction of bad observations that may cause an estimator to take on arbitrarily large aberrant values."

\item Dasgupta, Kumar, and Srikumar~\cite{AHW1996} write
``the breakdown point of an estimator is the largest fraction of the data that can be moved arbitrarily without perturbing the estimator to the boundary of the parameter space.''
\end{itemize}
\vspace{-.1in}
All of these definitions have similar meanings, and they are typically sufficient for the purpose of understanding a single estimator.
However, they are not mathematically rigorous, and it is difficult to use them to discuss the breakdown point of composite estimators.

\vspace{-.1in}
\paragraph{Composition of Estimators.}
In a bit more detail (we give formal definitions in Section \ref{A Formal Definition}), an estimator $E$ maps a data set to single value in another space, sometimes the same as a single data point.  For instance the mean or the median are simple estimators on one-dimensional data.
A composite $E_1$-$E_2$ estimator applies two estimators $E_1$ and $E_2$ on data stored in a hierarchy.  Let $\Eu{P} = \{P_1, P_2, \ldots, P_n\}$ be a set of subdata sets, where each subdata set $P_i = \{p_{i,1}, p_{i,2}, \ldots, p_{i,k}\}$ has individual data readings.  Then the $E_1$-$E_2$ estimator reports $E_2(E_1(P_1), E_1(P_2), \ldots, E_1(P_n))$, that is the estimator $E_2$ applied to the output of estimator $E_1$ on each subdata set.

\subsection{Examples of Estimator Composition}
\label{sec:examples}
Composite estimators arise in many scenarios in data analysis.

\vspace{-.1in}
\paragraph{Uncertain Data.}
For instance, in the last decade there has been increased focus on the study of uncertainty data~\cite{1644250,JLP11,CM08} where instead of analyzing a data set, we are given a model of the uncertainty of each data point.  Consider tracking the summarization of a group of $n$ people based on noisy GPS measurements.  For each person $i$ we might get $k$ readings of their location $P_i$, and use these $k$ readings as a discrete probability distribution of where that person might be.  Then in order to represent the center of this set of people a natural thing to do would be to estimate the location of each person as $x_i \leftarrow E_1(P_i)$, and then use these estimates to summarize the entire group $E_2(x_1, x_2, \ldots, x_n)$.
Using the \textsf{mean} as $E_1$ and $E_2$ would be easy, but would be susceptible to even a single outrageous outlier (all people are in Manhattan, but a spurious reading was at $(0,0)$ lat-long, off the coast of Africa).  An alternative is to use the $L_1$-median for $E_1$ and $E_2$, that is known to have an optimal breakdown point of $0.5$.  But what is the breakdown point of the $E_1$-$E_2$ estimator?

\vspace{-.1in}
\paragraph{Robust Analysis of Bursty Behavior.}
Understanding the robustness of estimators can also be critical towards how much one can ``game'' a system.
For instance, consider a start-up media website that gets bursts of traffic from memes they curate.  They publish a statistic showing the median of the top half of traffic days each month, and aggregate these by taking the median of such values over the top half of all months.  This is a composite estimator, and they proudly claim, even through they have bursty traffic, it is robust (each estimator has a breakdown point of $0.25$).  If this composite estimator shows large traffic, should a potential buyer of this website by impressed?  Is there a better, more robust estimator the potential buyer could request?
If the media website can stagger the release of its content, how should they distribute it to maximize this composite estimator?

\vspace{-.1in}
\paragraph{Part of the Data Analysis Pipeline.}
This process of estimator composition is very common in broad data analysis literature.  This arises from the idea of an ``analysis pipeline'' where at several stages estimators or analysis is performed on data, and then further estimators and analysis are performed downstream.
In many cases a robust estimator like the median is used, specifically for its robustness properties, but there is no analysis of how robust the composition of these estimators is.

\subsection{Main Results}

\vspace{-.05in}
This paper initiates the formal and general study of the robustness of composite estimators.  

%


\vspace{-.1in}
\begin{itemize}
\item  In Subsection \ref{A Formal Definition}, we give two formal definitions of breakdown points which are both required to prove composition theorem.  One variant of the definition closely aligns with other formalizations~\cite{Ham71,DG07}, while another is fundamentally different.

\item   The main result provides general conditions under which an $E_1$-$E_2$ estimator with breakdown points $\beta_1$ and $\beta_2$, has a breakdown point of $\beta_1\beta_2$ (Theorem \ref{theorem the upper bound of the breakdown point of E1-E2} in Subsection \ref{Definition of E1-E2 Estimator}).

\item Moreover, by showing examples where our conditions do not strictly apply, we gain an understanding of how to circumvent the above result.  An example is in composite percentile estimators (e.g., $E_1$ returns the $25$th percentile, and $E_2$ the $75$th percentile of a ranked set).  These composite estimators have larger breakdown point than $\beta_1 \cdot \beta_2$.

\item The main result can extended to multiple compositions, under suitable conditions, so for instance an $E_1$-$E_2$-$E_3$ estimator has a breakdown point of $\beta_1\beta_2\beta_3$ (Theorem \ref{theorem the breakdown point of E1-E2-E3} in Subsection \ref{sec:multi-comp}).
    This implies that long analysis chains can be very suspect to a few carefully places outliers since the breakdown point decays exponentially in the length of the analysis chain.

\item  In Section \ref{Application}, we highlight several applications of this theory, including robust regression, robustness of p-values, a depth-3 composition, and how to advantageously manipulate the observation about percentile estimator composition.
We demonstrate a few more applications with simulations in Section \ref{simulation}.

%

\end{itemize}

\section{Robustness of Estimator Composition}


\subsection{Formal Definitions of Breakdown Points}
\label{A Formal Definition}

%
%

%
%

\vspace{-.1in}
In this paper, we give two definitions for the breakdown point: \emph{Asymptotic Breakdown Point} and \emph{Asymptotic Onto-Breakdown Point}.  The first definition,
Asymptotic Breakdown Point, is similar to the classic formal definitions in  \cite{Ham71} and \cite{DG07} (including their highly technical nature), although their definitions of the estimator are slightly different leading to some minor differences in special cases.
However our second definition, Asymptotic Onto-Breakdown Point, is a structurally new definition, and we illustrate how it can result in significantly different values on some common and useful estimators.
Our main theorem will require both definitions, and the differences in performance will lead to several new applications and insights.

We define an \emph{estimator} $E$ as a function from the collection of some finite subsets of a metric space $(\mathscr{X},d)$ to another metric space  $(\mathscr{X}',d')$:
\begin{equation}  \label{def of estimator}
E:\ \mathscr{A}\subset\{X\subset\mathscr{X} \mid 0<|X|<\infty\}\mapsto\mathscr{X}',
\end{equation}
where $X$ is a multiset.  This means if $x\in X$ then $x$ can appear more than once in $X$, and the multiplicity of elements will be considered when we compute $|X|$.

\paragraph{Finite Sample Breakdown Point.}
For estimator $E$ defined in \eqref{def of estimator} and positive integer $n$ we define its \emph{finite sample breakdown point} $g_E(n)$ over a set $M$ as
\begin{equation}   \label{def of g_E(n)}
g_E(n)=\begin{cases} \max(M) & \text{ if } M \neq\emptyset\\
                   0   & \text{ if } M =\emptyset\\
        \end{cases}
\end{equation}
where for $\rho(x',X)=\max_{x\in X}d(x',x)$ is the distance from $x'$ to the furthest point in $X$,
\begin{equation}   \label{def of M}
\begin{split}
M=\{m \in [0,n] \mid \; & \forall X\in \mathscr{A}, |X|=n,
\forall\ G_1>0, \exists\  G_2=G_2(X,G_1) \text{ s.t. }
\\ &
\forall X'\in \mathscr{A},
 \text{ if } |X'|=n \text{ and }
|\{x'\in X' \mid \rho(x',X)>G_1\}|\leq m
\\ &
\text{ then }
d'(E(X),E(X'))\leq G_2\}.
\end{split}
\end{equation}
For an estimator $E$ in \eqref{def of estimator} and $X\in \mathscr{A}$, the finite sample breakdown point $g_E(n)$ means if the number of
unbounded points in $X'$ is at most $g_E(n)$, then $E(X')$ will be bounded.
Lets break this definition down a bit more.
The definition holds over all data sets $X \in \mathscr{A}$ of size $n$, and for all values $G_1 > 0$ and some value $G_2$ defined as a function $G_2(X,G_1)$ of the data set $X$ and value $G_1$.
Then $g_E(n)$ is the maximum value $m$ (over all $X$, $G_1$, and $G_2$ above) such that for all $X' \in \mathscr{A}$ with $|X'|=n$ then
$|\{x' \in X' \mid \rho(x',X) > G_1\}| \leq m$
(that is at most $m$ points are further than $G_1$ from $X$)
where the estimators are close, $d'(E(X),E(X')) \leq G_2$.

For example, consider a point set $X = \{0, 0.15, 0.2, 0.25, 0.4, 0.55, 0.6, 0.65, 0.72, 0.8, 1.0\}$ with $n = 11$ and median $0.55$.  If we set $G_1 = 3$, then we can consider sets $X'$ of size $11$ with fewer than $m$ points that are either greater than $3$ or less than $-2$.
 This means in $X'$ there are at most $m$ points which are greater than $3$ or less than $-2$, and all other $n-m$ points are in $[-2,3]$.
Under these conditions, we can (conservatively) set $G_2 = 4$, and know that for values of $m $ as $1, 2, 3, 4$, or $5$, then the median of $X'$ must be between $-3.45$ and $4.55$; and this holds no matter where we set those $m$ points (e.g., at $20$ or at $1000$).
This does not hold for $m \geq 6$, so $g_E(11) = 5$.

%
%

\vspace{-.1in}	
\paragraph{Asymptotic Breakdown Point.}
If the limit $\lim_{n\rightarrow\infty}\frac{g_E(n)}{n}$ exists, then we define this limit
\begin{equation} \label{asymptotic breakdown point}
\beta=\lim_{n\rightarrow \infty}\frac{g_E(n)}{n}
\end{equation}
 as the \emph{asymptotic breakdown point}, or \emph{breakdown point} for short, of the estimator $E$.

%

\theoremstyle{remark}
\newtheorem{Remark of first theorem}{Remark}
\begin{Remark of first theorem}
It is not hard to see that many common estimators satisfy the conditions.  For example, the median, $L_1$-median~\cite{Alo06}, and Siegel estimators~\cite{Sie82} all have asymptotic breakdown points of $0.5$.
\end{Remark of first theorem}

\vspace{-.1in}
\paragraph{Asymptotic Onto-Breakdown Point.}
For an estimator $E$ given in \eqref{def of estimator} and positive integer $n$, if
\begin{equation*} 
\begin{split}
\widetilde{M}=
\{0\leq m\leq n \mid & \; \forall\ X \in\mathscr{A},|X|=n,
\forall\ y\in \mathscr{X}',
\\ & \;
\exists\ X'\in \mathscr{A}
\text{ s.t. }
|X'|=n, |X\cap X'|=n-m, E(X')=y\}
\end{split}
\end{equation*}
is not empty, we define
\begin{equation} \label{def of f_E(n)}
f_E(n)= \min(\widetilde{M}).
\end{equation}
The definition of $f_E(n)$ implies, if we change $f_E(n)$ elements in $X$, we can make $E$ become \emph{any} value in $\mathscr{X}'$: it is onto.
In contrast $g_E(n)$ only requires $E(X')$ to become far from $E(X)$, perhaps only in one direction.
Then the \emph{asymptotic onto-breakdown point} is defined as the following limit if it exists
\begin{equation}   \label{onto breakdown point}
\lim_{n\rightarrow \infty}\frac{f_E(n)}{n}.
\end{equation}


\theoremstyle{remark}
\newtheorem{remark2}[Remark of first theorem]{Remark}
\begin{remark2}   \label{remark different breakdownpoint}
For a quantile estimator $E$ that returns a percentile other than the $50$th, then
$
\lim_{n\rightarrow \infty}\frac{g_E(n)}{n}
\neq
\lim_{n\rightarrow \infty}\frac{f_E(n)}{n}.
$
For instance, if $E$ returns the $25$th percentile of a ranked set, setting only $25\%$ of the data points to $-\infty$ causes $E$ to return $-\infty$; hence $\lim_{n\rightarrow \infty}\frac{g_E(n)}{n}  = 0.25$.   And while any value less than the original $25$th percentile can also be obtained; to return a value larger than the largest element in the original set, at least $75\% $ of the data must be modified, thus $\lim_{n\rightarrow \infty}\frac{f_E(n)}{n} = 0.75$.

As we will observe in Section \ref{Application}, this nuance in definition regarding percentile estimators will allow for some interesting composite estimator design.
\end{remark2}

\subsection{Definition of $E1$-$E2$ Estimators, and their Robustness}
\label{Definition of E1-E2 Estimator}

\vspace{-.1in}
We consider the following two estimators:
\begin{equation} \label{def of estimator E1}
E_1:\ \mathscr{A}_1\subset\{X\subset\mathscr{X}_1 \mid 0<|X|<\infty\}\mapsto\mathscr{X}_2 ,
\end{equation}
\vspace{-.2in}
\begin{equation} \label{def of estimator E2}
E_2:\ \mathscr{A}_2\subset\{X\subset\mathscr{X}_2 \mid 0<|X|<\infty\}\mapsto\mathscr{X}_2' ,
\end{equation}
where any finite subset of $E_1(\mathscr{A}_1)$, the range of $E_1$, belongs to $\mathscr{A}_2$.
Suppose $P_i \in \mathscr{A}_1$, $|P_i|=k$ for $i=1,2,\cdots,n$ and $P_\all=\uplus_{i=1}^nP_i$,
where $\uplus$ means if $x$ appears $n_1$ times in $X_1$ and $n_2$ times in $X_2$
then $x$ appears $n_1+n_2$ times in $X_1\uplus X_2$.
We define
\begin{equation} \label{def of estimator E}
E(P_\all)=E_2\left(E_1(P_1),E_1(P_2),\cdots,E_1(P_n)\right).
\end{equation}

%
%
%
\theoremstyle{theorem}
\newtheorem{theorem 1}{Theorem}
\begin{theorem 1}\label{theorem the lower bound of the breakdown point of E1-E2}
Suppose $g_{E_1}(k)$ and $g_{E_2}(n)$ are the finite sample breakdown points of estimators $E_1$ and $E_2$ which are given by \eqref{def of estimator E1} and \eqref{def of estimator E2} respectively. If $g_E(nk)$ is the finite sample breakdown point of $E$ given by \eqref{def of estimator E}, then we have
\begin{equation} \label{the lower bound of the breakdown point of E1-E2}
g_{E_2}(n)g_{E_1}(k)\leq g_E(nk).
\end{equation}
and if
\begin{equation*}
\beta_1=\lim_{k\rightarrow \infty}\frac{g_{E_1}(k)}{k},\ \ \beta_2=\lim_{n\rightarrow \infty}\frac{g_{E_2}(n)}{n},
\beta=\lim_{n,k\rightarrow \infty}\frac{g_{E}(nk)}{nk}
\end{equation*}
and all exist, then  \vspace{-.1in}
\begin{equation}   \label{lower boud of beta}
\beta_1\beta_2\leq\beta .
\end{equation}
\end{theorem 1}

\begin{proof}
For any fixed $G_1>0$, and any subsets
$
P_1',P_2',\cdots,P_n' \in \mathscr{A}_1
$
satisfying
$
|P_1'|=|P_2'|=\cdots=|P_n'|=k ,
$
and
\begin{equation} \label{|{p' in P_all'| rho(p',P_all)>G_1}|}
|\{p'\in P_\all'|\ \rho(p',P_\all)>G_1\}|\leq g_{E_2}(n)g_{E_1}(k)
\end{equation}
where $P_\all'=\uplus_{i=1}^nP_i'$, we introduce the notation
\begin{equation*}
\begin{split}
X=\{E_1(P_1),E_1(P_2),\cdots,E_1(P_n)\}, \ \ \
X'=\{E_1(P_1'),E_1(P_2'),\cdots,E_1(P_n')\}.
\end{split}
\end{equation*}
So, in order to prove \eqref{the lower bound of the breakdown point of E1-E2}, we only need to bound $E(P_\all')$.

We define
\begin{equation}  \label{def of I1}
I_1=\left\{1\leq i\leq n|\ |\{p'\in P_i'|\ \rho(p',P_i)>G_1\}|>g_{E_1}(k)\right\}
\end{equation}
and then have
\begin{equation}  \label{|I_1|leq g_{E_2}(n)}
|I_1|\leq g_{E_2}(n).
\end{equation}
Otherwise, since $\rho(p',P_i)>G_1$ implies $\rho(p',P_\all)>G_1$, from $|I_1|>g_{E_2}(n)$ and \eqref{def of I1}
we can obtain
\begin{equation*}
|\{p'\in P_\all'|\ \rho(p',P_\all)>G_1\}|> g_{E_2}(n)g_{E_1}(k)
\end{equation*}
which is contradictory to \eqref{|{p' in P_all'| rho(p',P_all)>G_1}|}.

For any $i\notin I_1$, we have
$
|\{p'\in P_i'|\ \rho(p',P_i)>G_1\}|\leq g_{E_1}(k),
$
so, from the definition of $g_{E_1}(k)$ we know
\begin{equation*}
\exists\ G_2^i=G_2^i(P_i, G_1),\ \text{s.t.}\  d_2(E_1(P_i'),E_1(P_i))\leq G_2^i\ \ \forall\ i\notin I_1.
\end{equation*}
where $d_2$ is the metric of space $\mathscr{X}_2$.
Let
\begin{equation*}
G_2=\max_{i\notin I_1}G_2^i+\max_{1\leq i,j\leq n}d_2(E_1(P_i), E_1(P_j))
\end{equation*}
then we have
\begin{equation} \label{rho(E_1(P_i'),X)leq G_2, forall i notin I_1}
\rho(E_1(P_i'),X)\leq G_2, \forall\ i\notin I_1.
\end{equation}
Defining
$
I_2=\{1\leq i \leq n \mid \rho(E_1(P_i'),X)>G_2\}
$
from \eqref{rho(E_1(P_i'),X)leq G_2, forall i notin I_1} we have $I_2\subset I_1$, which implies
$
|I_2|\leq |I_1|\leq g_{E_2}(n)
$
by \eqref{|I_1|leq g_{E_2}(n)}. Therefore, from the definition of $g_{E_2}(n)$, we have
\begin{equation*}
\begin{split}
\exists\ G_3=G_3(X,G_2) \text{ s.t. }
\|E(P_\all')-E(P_\all)\|=\|E_2(X')-E_2(X)\|\leq G_3,
\end{split}
\end{equation*}
which implies \eqref{the lower bound of the breakdown point of E1-E2}, and \eqref{lower boud of beta}
can be obtained from  \eqref{the lower bound of the breakdown point of E1-E2} directly. Thus, the proof is completed.
\end{proof}

\theoremstyle{remark}
\newtheorem{remark 3}[Remark of first theorem]{Remark}
\begin{remark 3}
Under the condition of Theorem \ref{theorem the lower bound of the breakdown point of E1-E2}, we cannot guarantee
$\beta=\beta_1\beta_2$. For example, suppose $E_1$ and $E_2$ take the 25th percentile and
the 75th percentile of a ranked set of real numbers respectively. So, we have $\beta_1=\beta_2=\frac{1}{4}$. However,
$\beta=\frac{1}{4}\cdot\frac{3}{4}=\frac{3}{16}$.

In fact, the limit of $\frac{g_E(nk)}{nk}$ as $n,k \rightarrow \infty$ may even not
exist. For example,  suppose $E_1$ takes the 25th percentile of a ranked set of real numbers. When $n$ is odd
$E_2$ takes the the 25th percentile of  a ranked set of $n$ real numbers, and when $n$ is even
$E_2$ takes the the 75th percentile of  a ranked set of $n$ real numbers. Thus, $\beta_1=\beta_2=\frac{1}{4}$, but
$g_E(nk)\approx \frac{1}{4}nk$ if $n$ is odd, and   $g_E(nk)\approx \frac{1}{4}\cdot\frac{3}{4}nk$ if $n$ is even, which implies
$\lim_{n,k\rightarrow \infty}\frac{g_E(nk)}{nk}$ does not exist.
\end{remark 3}


Therefore, to guarantee $\beta$ exist and $\beta=\beta_1\beta_2$, we introduce the definition of asymptotic onto-breakdown point in
\eqref{onto breakdown point}. As shown in \emph{Remark \ref{remark different breakdownpoint}}, the values of \eqref{asymptotic breakdown point}
and \eqref{onto breakdown point} may be not equal. However,  with the condition of the asymptotic breakdown point and asymptotic onto-breakdown point of $E_1$ being the same, we can finally state our desired clean result.


\theoremstyle{theorem}
\newtheorem{theorem 2}[theorem 1]{Theorem}
\begin{theorem 2}\label{theorem the upper bound of the breakdown point of E1-E2}
For estimators $E_1$, $E_2$ and $E$ given by  \eqref{def of estimator E1}, \eqref{def of estimator E2} and \eqref{def of estimator E} respectively,
suppose $g_{E_1}(k)$, $g_{E_2}(n)$ and $g_E(nk)$ are defined by \eqref{def of g_E(n)}, and $f_{E_1}(k)$ is defined by
\eqref{def of f_E(n)}. Moreover, $E_1$ is an onto function and for any fixed positive integer $n$ we have
\begin{equation}  \label{condition of E_2}
\begin{split}
&\exists\  X\in\mathscr{A}_2,|X|=n, G_1>0, \text{s.t. } \forall\  G_2>0, \exists\  X'\in \mathscr{A}_2 \text{ satisfying } \\
&|X'|=n, |X'\setminus X|=g_{E_2}(n)+1, \text{ and } d_2'(E_2(X),E_2(X'))>G_2 .
\end{split}
\end{equation}
where $d_2'$ is the metric of space $\mathscr{X}_2'$.

If
\begin{equation}   \label{beta1 and beta2 both exist 1}
\beta_1=\lim_{k\rightarrow \infty}\frac{g_{E_1}(k)}{k}=\lim_{k\rightarrow \infty}\frac{f_{E_1}(k)}{k},  \;\; \text{ and } \;\;
\beta_2=\lim_{n\rightarrow \infty}\frac{g_{E_2}(n)}{n}
\end{equation}
both exist, then
\begin{equation} \label{beta exists}
\beta=\lim_{n,k\rightarrow \infty}\frac{g_{E}(nk)}{nk}
\text{  exists } \;\;\;  \text{ and  } \; \; \;
\beta=\beta_1\beta_2.
\end{equation}
\end{theorem 2}

\begin{proof}
For any fixed positive integer $n$, we can find  $X=\{x_1,x_2,\cdots,x_n\}\in\mathscr{A}_2,$ and $G_1>0$ satisfying \eqref{condition of E_2}.
Since $E_1$ is an onto function, we can find  $P_\all=\uplus_{i=1}^nP_i$ such that $P_i\in \mathscr{A}_1$ and $E_1(P_i)=x_i$ for all $1\leq i\leq n$.

From \eqref{condition of E_2}, we know for any $G_2>0$, we can find $X'\in\mathscr{A}_2$ such that $|X'|=n$, $|X'\setminus X|=g_{E_2}(n)+1$ and
\begin{equation*}
d'(E_2(X),E_2(X'))>G_2.
\end{equation*}
 This implies the number of different elements between $X$ and $X'$ is $g_{E_2}(n)+1$. For any $x_i'\in X'\setminus X$, we can find $P_i'\in \mathscr{A}_1$ such that $|P_i'|=k$, $E_1(P_i')=x_i'$ and $|P_i'\setminus P_i|=f_{E_1}(k)$. So, we only need to change
$f_{E_1}(k) (g_{E_2}(n)+1)$ points of $P_\all$, and then we can obtain $P_\all'$ such that $|P_\all' \setminus P_\all|=f_{E_1}(k) (g_{E_2}(n)+1)$
and $d'(E(P_\all),E(P_\all'))>G_2$. This implies
\begin{equation} \label{g_E(nk)leq f_{E_1}(k) (g_{E_2}(n)+1)}
g_E(nk)\leq f_{E_1}(k) (g_{E_2}(n)+1).
\end{equation}
Therefore, from Theorem \ref{theorem the lower bound of the breakdown point of E1-E2} and \eqref{g_E(nk)leq f_{E_1}(k) (g_{E_2}(n)+1)}
we have
\begin{equation} \label{frac{g_{E_1}(k)}{k} frac{g_{E_2}(n)}{n}leq frac{g_E(nk)}{nk}leq frac{f_{E_1}(k)}{k} frac{(g_{E_2}(n)+1)}{n}}
\frac{g_{E_1}(k)}{k} \frac{g_{E_2}(n)}{n}\leq \frac{g_E(nk)}{nk}\leq \frac{f_{E_1}(k)}{k} \frac{(g_{E_2}(n)+1)}{n}.
\end{equation}
Letting $n$ and $k$ go to infinity in \eqref{frac{g_{E_1}(k)}{k} frac{g_{E_2}(n)}{n}leq frac{g_E(nk)}{nk}leq frac{f_{E_1}(k)}{k} frac{(g_{E_2}(n)+1)}{n}}, we obtain  \eqref{beta exists} from  \eqref{beta1 and beta2 both exist 1}.
Thus, the proof of this theorem is completed.
\end{proof}

\theoremstyle{remark}
\newtheorem{remark3}[Remark of first theorem]{Remark}
\begin{remark3}

Without the introduction of $f_E(n)$, we cannot even guarantee $\beta\leq \beta_1$ or $\beta\leq \beta_2$ only under the condition of Theorem \ref{theorem the lower bound of the breakdown point of E1-E2}, even if $E_1$ and $E_2$ are both onto functions. For example, for any
$P=\{p_1,p_2,\cdots,p_k\}\subset \mathbb{R}$ and $X=\{x_1,x_2,\cdots,x_n\}\subset \mathbb{R}$, we define
$E_1(P) = 1/\text{median}(P)$ (if $\text{median}(P) \neq 0$, otherwise define $E_1(P) = 0$)
and
$E_2(X)=\text{median}(y_1,y_2,\cdots,y_n),
$
where $y_i$ $(1\leq y \leq n)$ is given by $y_i = 1/x_i$ (if $x_i \neq 0$, otherwise define $y_i = 0$).
Since $g_{E_1}(k)=g_{E_2}(n)=0$ for all $n,k$, we have $\beta_1=\beta_2=0$. However, in order to make
$
E_2(E_1(P_1),E_1(P_2),\cdots, E_1(P_n))\rightarrow +\infty,
$
we need to make about $\frac{n}{2}$ elements in
$
\{E(P_1),E(P_2),\cdots,E(P_n)\}
$
go to $0+$. To make
$E_1(P_i)\rightarrow 0+$, we need to make about $\frac{k}{2}$ points in $P_i$ go to $+\infty$. Therefore, we have $g_E(nk)\approx\frac{n}{2}\cdot \frac{k}{2}$ and $\beta=\frac{1}{4}$.
\end{remark3}

\subsection{Multi-level Composition of Estimators}
\label{sec:multi-comp}

\vspace{-.1in}
To study the breakdown point of composite estimators with more than two levels, we introduce the following estimator:
\begin{equation} \label{def of estimator E3}
E_3:\ \mathscr{A}_3\subset\{X\subset\mathscr{X}_2' \mid 0<|X|<\infty\}\mapsto\mathscr{X}_3' ,
\end{equation}
where any finite subset of $E_2(\mathscr{A}_2)$, the range of $E_2$, belongs to $\mathscr{A}_3$.
Suppose $P_{i,j} \in \mathscr{A}_1$, $|P_{i,j}|=k$ for $i=1,2,\cdots,n$, $j=1,2,\cdots,m$  and $P_\all^j=\uplus_{i=1}^nP_{i,j}$,
$P_\all=\uplus_{j=1}^mP_\all^j$. We define
\begin{equation} \label{def of estimator E 3 level}
E(P_\all)=E_3\left(E_2(\widetilde{P}_\all^1),E_2(\widetilde{P}_\all^2),\cdots,E_2(\widetilde{P}_\all^m)\right),
\end{equation}
where $\widetilde{P}_\all^j=\{E_1(P_{1,j}),E_1(P_{2,j}),\cdots,E_1(P_{n,j})\}$, for $j=1,2,\cdots,m$.

From Theorem \ref{theorem the upper bound of the breakdown point of E1-E2}, we can obtain the following theorem about the breakdown point of $E$ in \eqref{def of estimator E 3 level}.


\theoremstyle{theorem}
\newtheorem{theorem 3}[theorem 1]{Theorem}
\begin{theorem 3}\label{theorem the breakdown point of E1-E2-E3}
For estimators $E_1$, $E_2$, $E_3$ and $E$ given by  \eqref{def of estimator E1}, \eqref{def of estimator E2}, \eqref{def of estimator E3} and \eqref{def of estimator E 3 level} respectively,
suppose $g_{E_1}(k)$, $g_{E_2}(n)$,  $g_{E_3}(m)$ and $g_E(mnk)$ are defined by \eqref{def of g_E(n)}, and $f_{E_1}(k)$, $f_{E_2}(n)$ are defined by
\eqref{def of f_E(n)}. Moreover, $E_1$ and $E_2$ are both onto functions, and for any fixed positive integer $m$ we have
\begin{equation*}  
\begin{split}
&\exists\  X\in\mathscr{A}_3,|X|=m, G_1>0, \text{s.t. } \forall\  G_2>0,
\exists\  X'\in \mathscr{A}_3 \\
& \text{ satisfying }
|X'|=m, |X'\setminus X|=g_{E_3}(m)+1,
\text{ and } d_3'(E_3(X),E_3(X'))>G_2 .
\end{split}
\end{equation*}
where $d_3'$ is the metric of space $\mathscr{X}_3'$.
If
\begin{equation}   \label{beta1 and beta2 both exist}
\begin{split}
\beta_1=\lim_{k\rightarrow \infty}\frac{g_{E_1}(k)}{k}=\lim_{k\rightarrow \infty}\frac{f_{E_1}(k)}{k}, \; \; \;
\beta_2=\lim_{n\rightarrow \infty}\frac{g_{E_2}(n)}{n}=\lim_{n\rightarrow \infty}\frac{f_{E_2}(n)}{n},
\end{split}
\end{equation}
and
$
\beta_3=\lim_{m\rightarrow \infty}\frac{g_{E_3}(m)}{m}
$
all exist, then
\begin{equation} \label{beta exists three levels}
\beta=\lim_{m,n,k\rightarrow \infty}\frac{g_{E}(mnk)}{mnk}
\textrm{ exist } \;\;\; \text{ and } \;\;\;
\beta=\beta_1\beta_2\beta_3.
\end{equation}

\end{theorem 3}

\begin{proof}
We define an estimator $\widetilde{E}$:
\begin{equation*}
\widetilde{E}(\widetilde{P}_\all^j)=E_2(E_1(P_{1,j}),E_1(P_{2,j}),\cdots,E_1(P_{n,j}))
\end{equation*}
for $j=1,2,\cdots,m$, and first prove the breakdown point of $\widetilde{E}$ is
$\tilde{\beta}=\beta_1\beta_2.$

For any fixed $y\in \mathscr{X}_2'$ and $X=\{E_1(P_1),E_1(P_2),\cdots,E_1(P_n)\}$, we can find
 $X'\in \mathscr{A}_2$ such that $|X'|=n$, $|X\cap X'|=n- f_{E_2}(n)$ and $E_2(X')=y$. For any element $y'\in X'\setminus (X\cap X')$,
 we can find $E_1(P_i)\in X\setminus(X\cap X')$ and $P_i'\in \mathscr{A}_1$ such that $|P_i'|=k$, $|P_i\cap P_i'|=k-g_{E_1}(k)$ and $E_1(P_i')=y'$. This implies we can find a set $P_\all' \subset\mathscr{X}_1$ such that $|P_\all'|=nk$, $|P_\all\cap P_\all'|=nk-f_{E_2}(n)f_{E_1}(k)$
and $\widetilde{E}(P_\all')=y$, i.e. we only need to change $f_{E_2}(n)f_{E_1}(k)$ points in $P_\all$, and $\widetilde{E}$ can become any value. So, we have
\begin{equation}  \label{f_E(nk)leq f_{E_2}(n)f_{E_1}(k)}
f_{\widetilde{E}}(nk)\leq f_{E_2}(n)f_{E_1}(k).
\end{equation}


Applying Theorem \ref{theorem the lower bound of the breakdown point of E1-E2} to $E_1$ and $E_2$, we obtain
\begin{equation} \label{the lower bound of the breakdown point of E1-E2 three level}
g_{E_2}(n)g_{E_1}(k)\leq g_{\widetilde{E}}(nk).
\end{equation}
Since $g_{\widetilde{E}}(nk)<f_{\widetilde{E}}(nk)$, from \eqref{f_E(nk)leq f_{E_2}(n)f_{E_1}(k)} and \eqref{the lower bound of the breakdown point of E1-E2 three level},
we have
\begin{equation} \label{gE2(n)gE1(k)<fE2(n)fE1(k)}
\frac{g_{E_2}(n)}{n}\frac{g_{E_1}(k)}{k}\leq \frac{g_{\widetilde{E}}(nk)}{nk}<\frac{f_{\widetilde{E}}(nk)}{nk}\leq \frac{f_{E_2}(n)}{n}\frac{f_{E_1}(k)}{k}.
\end{equation}
Letting $n,k$ go to infinity in \eqref{gE2(n)gE1(k)<fE2(n)fE1(k)}, from \eqref{beta1 and beta2 both exist}
we obtain the breakdown point of $\widetilde{E}$ is
\begin{equation*} 
\tilde{\beta}=\lim_{n,k\rightarrow \infty}\frac{g_{\widetilde{E}}(nk)}{nk}=\lim_{n,k\rightarrow \infty}\frac{f_{\widetilde{E}}(nk)}{nk}
=\beta_1\beta_2.
\end{equation*}

Since
$
E(P_\all)=E_3(\widetilde{E}(\widetilde{P}_\all^1),\widetilde{E}(\widetilde{P}_\all^2),\cdots,\widetilde{E}(\widetilde{P}_\all^m)),
$
we apply  Theorem \ref{theorem the upper bound of the breakdown point of E1-E2} to $\widetilde{E}$ and $E_3$, and then obtain
\eqref{beta exists three levels}. 
\end{proof}


\section{Applications}  \label{Application}

We next discuss several applications of our main theorems and observations.
Applications 2 and 4 are direct applications of the easy to use theorems.
Applications 1 and 3 take advantage of some of the nuances in definition, in particular the unexpected robustness of composing quantile estimators.

\subsection{Application 1 : Balancing Percentiles}

\vspace{-.1in}
For $n$ companies, for simplicity, assume each company has $k$ employees.
We are interested in the income of the regular employees of all companies, not the executives who may have exorbitant pay.
Let $p_{i,j}$ represents the income of the $j$th employee in the $i$th company.
Set $P_\all=\uplus_{i=1}^n P_i$ where the $i$th company has a set $P_i=\{p_{i,1},p_{i,2},\cdots,p_{i,k}\} \subset \mathbb{R}$ and for notational convenience $p_{i,1}\leq p_{i,2}\leq \cdots \leq p_{i,k}$ for $i\in\{1,2,\cdots,n\}$.
Suppose the income data $P_i$ of each company is preprocessed by a $45$-percentile estimator $E_1$  (median of lowest 90\% of incomes), with breakdown point $\beta_1 = 0.45$.
In theory $E_1(P_i)$ can better reflect the income of regular employees in a company, since there may be about $10\%$ of employees in the management of a company and their incomes are usually much higher than that of common employees.
So, the preprocessed data is
$
X=\{E_1(P_1),E_1(P_2),\cdots,E_1(P_n)\}.
$

If we define $E_2(X)=\text{median}(X)$ and $E(P_\all)=E_2(X)$, then the breakdown point of $E_2$ is $\beta_2=0.5$, and the breakdown points of $E$ is $\beta=\beta_1\beta_2=0.225$.

However, if we use another $E_2$, then $E$ can be more robust. For example, for $X=\{x_1,x_2,\cdots, x_n\}$ where $x_1\leq x_2\leq \cdots \leq x_n $, we can define $E_2$ as the $55$-percentile estimator (median of largest $90\%$ of incomes).
In order to make $E(P_\all)=E_2(X)=E_2(E_1(P_1),E_1(P_2),\cdots,E_1(P_n))$ go to infinity, we need to either move $55\%$ points of $X$ to $-\infty$  or move $45\%$ points of $X$ to $+\infty$. In either case, we need to move about $0.45\cdot0.55 nk$ points of $P_\all$ to infinity. This means the breakdown point of $E$ is $\beta=0.45\cdot0.55=0.2475$  which is greater than $0.225$.

This example implies if we know how the raw data is preprocessed by estimator $E_1$, we can choose a proper estimator $E_2$ to make the $E_1$-$E_2$ estimator more robust.

\subsection{Application 2 : Regression of $L_1$ Medians}

Suppose we want to use linear regression to robustly predict the weight of a person from his or her height, and we have multiple readings of each person's height and weight.
The raw data is $P_\all=\uplus_{i=1}^n P_i$ where for the $i$th person we have a set $P_i=\{p_{i,1},p_{i,2},\cdots,p_{i,k}\} \subset \mathbb{R}^2$ and $p_{i,j}=(x_{i,j},y_{i,j})$ for $i\in\{1,2,\cdots,n\},j\in\{1,2,\cdots,k\}$.
Here, $x_{i,j}$ and $y_{i,j}$ are the height and weight respectively of the $i$th person in their $j$th measurement.

One ``robust'' way to process this data, is to first pre-process each $P_i$ with its $L_1$-median~\cite{Alo06}:  $(\bar x_i, \bar y_i)  \leftarrow E_1(P_i)$, where $E_1(P_i) = \text{$L_1$-median}(P_i)$ has breakdown point $\beta_1 = 0.5$.
Then we could generate a linear model to predict weight $\hat y_i = a x+ b$ from the Siegel Estimator~\cite{Sie82}: $E_2(Z) = (a,b)$, with breakdown point $\beta_2 = 0.5$.
From Theorem \ref{theorem the upper bound of the breakdown point of E1-E2}
we immediately know the breakdown point of $E(P_\all)=E_2(E_1(P_1),E_1(P_2),\cdots,E_1(P_n))$ is $\beta=\beta_1\beta_2=0.5\cdot0.5=0.25$.

%
%

Alternatively, taking the Siegel estimator of $P_\all$ (i.e., returning $E_2(P_\all)$) would have a much larger breakdown point of $0.5$.  So a seemingly harmless operation of normalizing the data with a robust estimator (with optimal $0.5$ breakdown point) drastically decreases the robustness of the process.

\subsection{Application 3 : Significance Thresholds}

Suppose we are studying the distribution of the wingspread of fruit flies.
There are $n=500$ flies, and the variance of the true wingspread among these flies is on the order of $0.1$ units.  Our goal is to estimate the $0.05$ significance level of this distribution of wingspread among normal flies.

To obtain a measured value of the wingspread of the $i$th fly, denoted $F_i$,
we measure the wingspread of $i$th fly $k=100$ times independently, and obtain the measurement set $P_i=\{p_{i,1},p_{i,2},\cdots,p_{i,k}\}$.
The measurement is carried out by a machine automatically and quickly, which implies
the variance of each $P_i$ is typically very small, perhaps only $0.0001$ units, but there are outliers in $P_i$ with small chance due to possible
machine malfunction.  This malfunction may be correlated to individual flies because of anatomical issues, or it may have autocorrelation (the machine jams for a series of consecutive measurements).

To perform hypothesis testing we desire the $0.05$ significance level, so we are interested in the $95$th percentile of the set $F=\{F_1,F_2,\cdots,F_n\}$.
So a post processing estimator $E_2$ returns the 95th percentile of $F$ and has a breakdown point of $\beta_2 = 0.05$~\cite{HSP90}.
Now, we need to design an estimator $E_1$ to process the raw data $P_\all=\uplus_{i=1}^n P_i$  to obtain  $F=\{F_1,F_2,\cdots,F_n\}$. For example, we can define $E_1$ as
$
F_i= E_1(P_i)=\text{median}(P_i)
$
and estimator $E$ as
$
E(P_\all)=E_2(E_1(P_1),E_1(P_2),\cdots,E_1(P_n)).
$

Then, the breakdown point of $E_1$ is 0.5. Since the breakdown point of $E_2$ is 0.05, the breakdown point of the composite estimator
$E$ is $\beta=\beta_1\beta_2=0.5\cdot0.05=0.025$.  This means if the measurement machine malfunctioned only $2.5\%$ of the time, we could have an anomalous significant level, leading to false discovery.  Can we make this process more robust by adjusting $E_1$?

Actually, \emph{yes!}, we can use another pre-processing estimator to get a more robust $E$. Since the variance of each $P_i$ is only $0.0001$, we can let
$E_1$ return the 5th percentile of a ranked set of real numbers, then there is not much difference between $E_1(P_i)$ and the median of $P_i$.
(Note: this introduces a small amount of bias that can likely be accounted for in other ways.)
In order to make
$E(P_\all)=E_2(F)$ go to infinity we need to move $5\%$ points of
$X$ to $-\infty$ (causing $E_2$ to give an anomalous value) or $95\%$ points of $X$ to $+\infty$ (causing many, $95\%$, of the $E_1$ values, to give anomalous values).
In either case, we need to move about $5\%\cdot95\%$ points of $P_\all$ to infinity. So, the breakdown points of $E$ is $\beta=0.05\cdot0.95=0.0475$
which is greater than $0.025$.  That is, we can now sustain up to $4.75\%$ of the measurement machine's reading to be anomalous, almost double than before, without leading to an anomalous significance threshold value.

This example implies if we know the post-processing estimator $E_2$, we can choose a proper method to preprocess the raw data to make the $E_1$-$E_2$ estimator more robust.


\vspace{.1in}

\theoremstyle{remark}
\newtheorem{remark4}[Remark of first theorem]{Remark}
\begin{remark4}
A further study would be required to use such a composite estimator in practice due some bias it introduces.  To replicate the normalization process on new experimental data (e.g., on a new species with hypothesized long wingspread), we would probably need to make one of the following adjustments to the standard process of measuring the wingspread of the new species and directly comparing it to the significance threshold.
 (a) Also consider the $5$th percentile of the experimental measurements (with breakdown point $0.05$ instead of $0.5$).
 (b) Adjust the significance level by roughly $0.0001$ units (the variance over $P_i$) making it conservative with respect to the $5$th percentile versus the $50$th percentile decision of each fly's measurements, so the $50$th percentile could be used on the new experimental data.
 Or, (c) use a different percentile (say the $(95+\eps)$th percentile instead of $95$th) to balance the bias in using the $5$th percentile of measurements.
In the specific scenario we describe, we believe option (b) may be a very acceptable option with little lack in precision (due to difference in variance $0.1$ and $0.0001$) but with large gain in robustness.
\end{remark4}



\subsection{Application 4 : 3-Level Composition}

Suppose we want to use a single value to represent the temperature of the US in a certain day. There are $m=50$ states in the
country. Suppose each state has $n=100$ meteorological stations, and the station $i$ in state $j$ measures the local temperature $k=24$ times
to get the data $P_{i,j}=\{t_{i,j,1}, t_{i,j,2},\cdots, t_{i,j,k}\}$.
We define $P_\all^j=\uplus_{i=1}^n P_{i,j}$, $P_\all=\uplus_{j=1}^m P_\all^j$ and
\begin{align*}
E_1(P_{i,j}) &= \text{median}(P_{i,j}),\ \ \
E_2(P_\all^j) = \text{median}\left(E_1(P_{1,j}),E_1(P_{1,j}),\cdots,E_1(P_{n,j})\right)
\\
E(P_\all) &= E_3(E_2(P_\all^1),E_2(P_\all^2),\cdots,E_2(P_\all^m))
= \text{median}(E_2(P_\all^1),E_2(P_\all^2),\cdots,E_2(P_\all^m)).
\end{align*}
So, the break down points of $E_1$, $E_2$ and $E_3$ are $\beta_1=\beta_2=\beta_3=0.5$.
From Theorem \ref{theorem the breakdown point of E1-E2-E3}, we know the break down
point of $E$ is $\beta=\beta_1\beta_2\beta_3=0.125$.
Therefore, we know the estimator $E$ is not very robust, and it may be not a good choice to use $E(P_\all)$ to represent the temperature of the US in a certain day.

This example illustrates how the more times the raw data is aggregated, the more unreliable the final result can become.

\section{Simulations}  \label{simulation}

We next describe a few more scenarios where our new theory on estimator composition is relevant.  For these we simulate a couple of data sets to demonstrate how one might construct interesting algorithms from these ideas.

\subsection{Simulation 1 : Estimator Manipulation}

In this simulation we actually construct a method to relocate an estimator by modifying the smallest number of points possible.  We specifically target the $L_1$-median of $L_1$-medians since its somewhat non-trivial to solve for the new location of data points.

In particular, given a target point $p_0\in\mathbb{R}^2$ and a set of $nk$ points $P_\all=\uplus_{i=1}^nP_i$, where $P_i=\{p_{i,1},p_{i,2},\cdots,p_{i,k}\} \subset \mathbb{R}^2$, we use simulation to show that we only need to change $\tilde{n}\tilde{k}$ points of $P_\all$, then we can  get a new set $\widetilde{P}_\all=\uplus_{i=1}^n\widetilde{P}_i$ such that $\text{median}(\text{median}(\widetilde{P}_1),\text{median}(\widetilde{P}_2),\cdots,\text{median}(\widetilde{P}_n))=p_0$. Here, the "median" means $L_1$-median, and
\begin{equation*}
\tilde{n}=\begin{cases} \frac{1}{2} n & \text{ if $n$ is even} \\
                       \frac{1}{2}(n+1) & \text{ if $n$ is odd}
          \end{cases}\ , \ \
\tilde{k}=\begin{cases} \frac{1}{2} k & \text{ if $k$ is even} \\
                       \frac{1}{2}(k+1) & \text{ if $k$ is odd}
                       \end{cases}\ .
\end{equation*}
To do this, we first show that, given $k$ points $S=\{(x_i,y_i) \mid 1\leq i\leq k\}$ in $\mathbb{R}^2$, and a target point $(x_0,y_0)$, we can
change $\tilde{k}$ points of $S$ to make $(x_0,y_0)$ as the $L_1$-median of the new set.
As $n$ and $k$ grow, then $\tilde n \tilde k / (nk) = 0.25$ is the asymptotic breakdown point of this estimator, as a consequence of Theorem \ref{theorem the upper bound of the breakdown point of E1-E2}, and thus we may need to move this many points to get the result.

If $(x_0,y_0)$ is the $L_1$-median of the set $\{(x_i,y_i) \mid 1\leq i\leq k\}$, then we have~\cite{EWFP2009}: 
\begin{equation}  \label{the condition of L1 median}
\begin{split}
&\sum_{i=1}^k \frac{x_i-x_0}{\sqrt{(x_i-x_0)^2+(y_i-y_0)^2}}=0, \ \
\sum_{i=1}^k \frac{y_i-y_0}{\sqrt{(x_i-x_0)^2+(y_i-y_0)^2}}=0.
\end{split}
\end{equation}

We define $\vec{x}=(x_1,x_2,\cdots,x_{\tilde{k}})$, $\vec{y}=(y_1,y_2,\cdots,y_{\tilde{k}})$ and
\begin{small}
\begin{equation*}
\begin{split}
h(\vec{x},\vec{y})=
\left(\sum_{i=1}^k \frac{x_i-x_0}{\sqrt{(x_i-x_0)^2+(y_i-y_0)^2}}\right)^2
+\left(\sum_{i=1}^k \frac{y_i-y_0}{\sqrt{(x_i-x_0)^2+(y_i-y_0)^2}}\right)^2.
\end{split}
\end{equation*}
\end{small}

Since \eqref{the condition of L1 median} is the sufficient and necessary condition
for $L_1$-median, if we can find $\vec{x}$ and $\vec{y}$ such that $h(\vec{x},\vec{y})=0$, then $(x_0,y_0)$ is the
$L_1$-median of the new set.

Since
\begin{small}
\begin{equation*}
\begin{split}
\partial_{x_i}h(\vec{x},\vec{y})=&
2\Big(\sum_{j=1}^k \frac{x_j-x_0}{\sqrt{(x_j-x_0)^2+(y_j-y_0)^2}}\Big)\frac{(y_i-y_0)^2}{\big((x_i-x_0)^2+(y_i-y_0)^2\big)^{\frac{3}{2}}}\\
&-2\Big(\sum_{j=1}^k \frac{y_j-y_0}{\sqrt{(x_j-x_0)^2+(y_j-y_0)^2}}\Big)\frac{(x_i-x_0)(y_i-y_0)}{\big((x_i-x_0)^2+(y_i-y_0)^2\big)^{\frac{3}{2}}},
\end{split}
\end{equation*}
\begin{equation*}
\begin{split}
\partial_{y_i}h(\vec{x},\vec{y})=
&-2\Big(\sum_{j=1}^k \frac{x_j-x_0}{\sqrt{(x_j-x_0)^2+(y_j-y_0)^2}}\Big)\frac{(x_i-x_0)(y_i-y_0)}{\big((x_i-x_0)^2+(y_i-y_0)^2\big)^{\frac{3}{2}}}\\
&+2\Big(\sum_{j=1}^k \frac{y_j-y_0}{\sqrt{(x_j-x_0)^2+(y_j-y_0)^2}}\Big)\frac{(x_i-x_0)^2}{\big((x_i-x_0)^2+(y_i-y_0)^2\big)^{\frac{3}{2}}},
\end{split}
\end{equation*}
\end{small}
we can use gradient descent to compute $\vec{x}, \vec{y}$ to minimize $h$.
For the input $S=\{(x_i,y_i)|1\leq i\leq k\}$, we choose the initial value $\vec{x}_0=\{x_1,x_2,\cdots,x_{\tilde{k}}\}$,
$\vec{y}_0=\{y_1,y_2,\cdots,y_{\tilde{k}}\}$,
and then update $\vec{x}$ and $\vec{y}$ along the negative gradient direction of $h$, until the Euclidean norm of gradient is less than 0.00001.

The algorithm framework is then as follows, using the above gradient descent formulation at each step.  We first compute the $L_1$-median $m_i$ for each $P_i$, and then change $\tilde{n}$ points in $\{m_1,m_2,\cdots,m_n\}$ to obtain
\begin{equation*}
\{m_1',m_2',\cdots,m_{\tilde{n}}',m_{\tilde{n}+1},\cdots,m_n\}
\end{equation*}
 such that median$(m_1',m_2',\cdots,m_{\tilde{n}}',m_{\tilde{n}+1},\cdots,m_n)=p_0$.
For each $m_i'$, we change $\tilde{k}$ points in $P_i$ to obtain
\begin{equation*}
\widetilde{P}_i=\{p_{i,1}',p_{i,2}',\cdots,p_{i,\tilde{k}}', p_{i,\tilde{k}+1}, \cdots, p_{i,k}\}
\end{equation*}
such that median$(\widetilde{P}_i)=m_i'$.
Thus, we have
\begin{equation}  \label{the median of medians, new set}
\begin{split}
\text{median}\big(&\text{median}(\widetilde{P}_1),\cdots,\text{median}(\widetilde{P}_{\tilde{n}}),
\text{median}(P_{\tilde{n}+1}), \cdots, \text{median}(P_n)\big)=p_0.
\end{split}
\end{equation}

To show a simulation of this process, we use a uniform distribution to randomly generate $nk$ points in the region $[-10,10]\times[-10,10]$, and generate a target point $p_0=(x_0,y_0)$
in the region $[-20,20]\times[-20,20]$, and then use our algorithm to change $\tilde{n}\tilde{k}$ points in the given set, to make the new
set satisfy \eqref{the median of medians, new set}.
Table \ref{simulation 1} shows the result of running this experiment  for different $n$ and $k$, where
$(x_0',y_0')$ is the median of medians for the new set obtained by our algorithm.  It lists the various values $n$ and $k$, the corresponding values $\tilde n$ and $\tilde k$ of points modified, and the target point and result of our algorithm.
If we reduce the terminating condition, which  means
increasing the number of iteration, we can obtain a more accurate result, but only requiring the Euclidean norm of gradient to be less than
0.00001, we get very accurate results, within about $0.01$ in each coordinate.


\begin{table*}
\caption{The running result of Simulation 1.} \label{simulation 1}
\centering
\begin{tabular}{|c|c|c|c|l|l|}
\hline
$\ n\ $  & $\ k\ $ & $\ \tilde{n}\ $  & $\ \tilde{k}\ $  & $(x_0,y_0)$  & $(x_0',y_0')$ \\
 \hline
5  &  8  & 3 & 4& (0.9961,   1.0126) & (0.9961,  1.0126)\\
\hline
5  &  8  & 3 & 4& (10.7631, 11.0663) & (10.7025   11.0623)\\
\hline
10& 5&5&3  &  (-13.8252,   -4.7462) &(-13.8330,   -4.7482)\\
\hline
50 & 20 & 25 &10&  ( -14.7196,  -13.6728)&  (-14.7263,  -13.6784)\\
\hline
100  &  50  & 50  & 25 &( -14.0778,   18.3665)  &( -14.0773,   18.3658)\\
\hline
500 & 100 & 250 & 50 &(-15.8408,   -6.4259) &  (-15.8385,   -6.4250)\\
\hline
1000 & 200& 500 &100 & (18.6351,  -12.1014) & (18.7886,  -12.2011)\\
 \hline
\end{tabular}
\end{table*}

\begin{figure*}[h]
  \centering
  \includegraphics[width=0.8\textwidth]{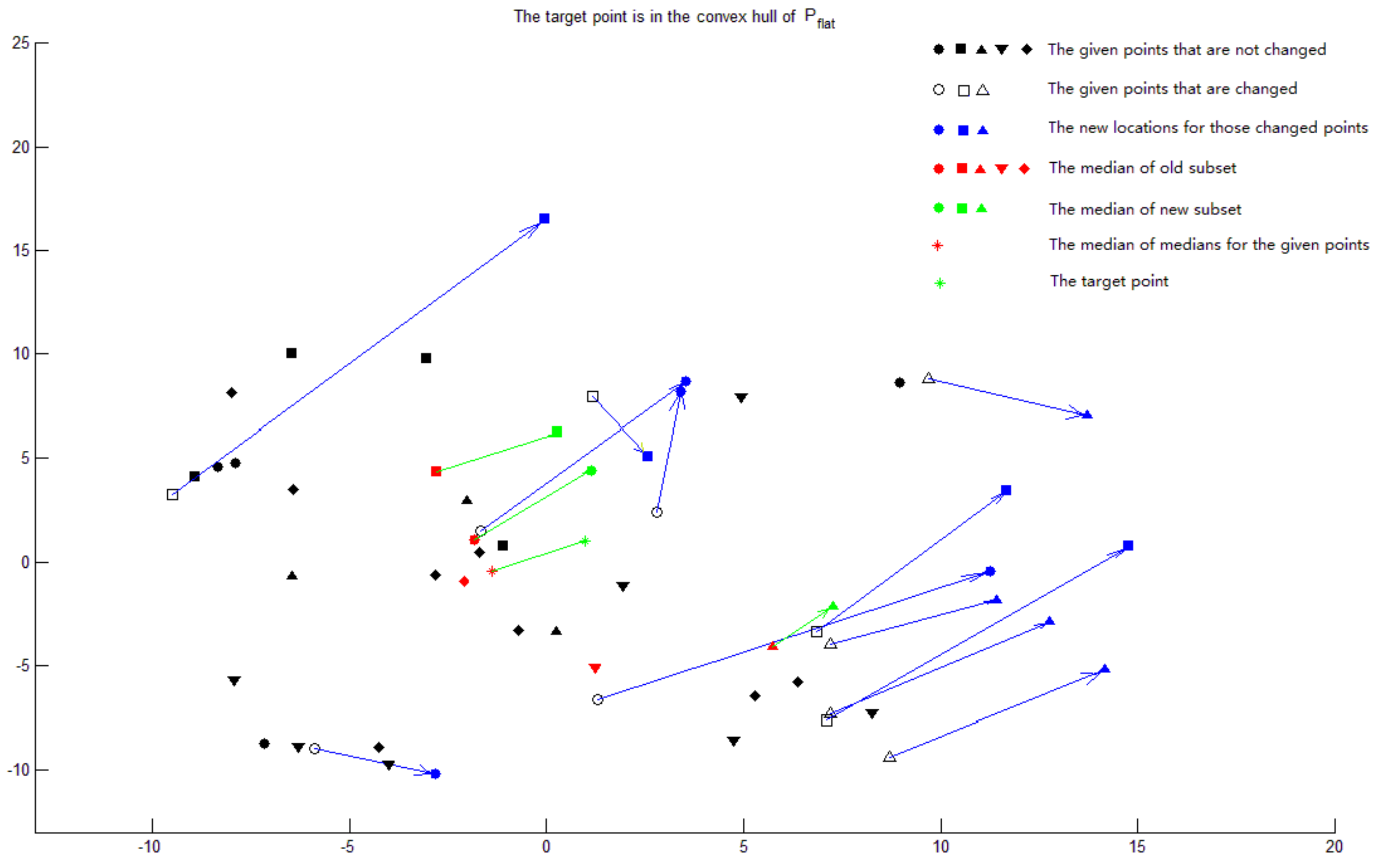}\\
  \caption{The running result for the case $n=5$, $k=8$, $(x_0,y_0)=(0.9961, 1.0126)$  in Table \ref{simulation 1}.}\label{fig1}
\end{figure*}

\begin{figure*}[h]
  \centering
  \includegraphics[width=0.8\textwidth]{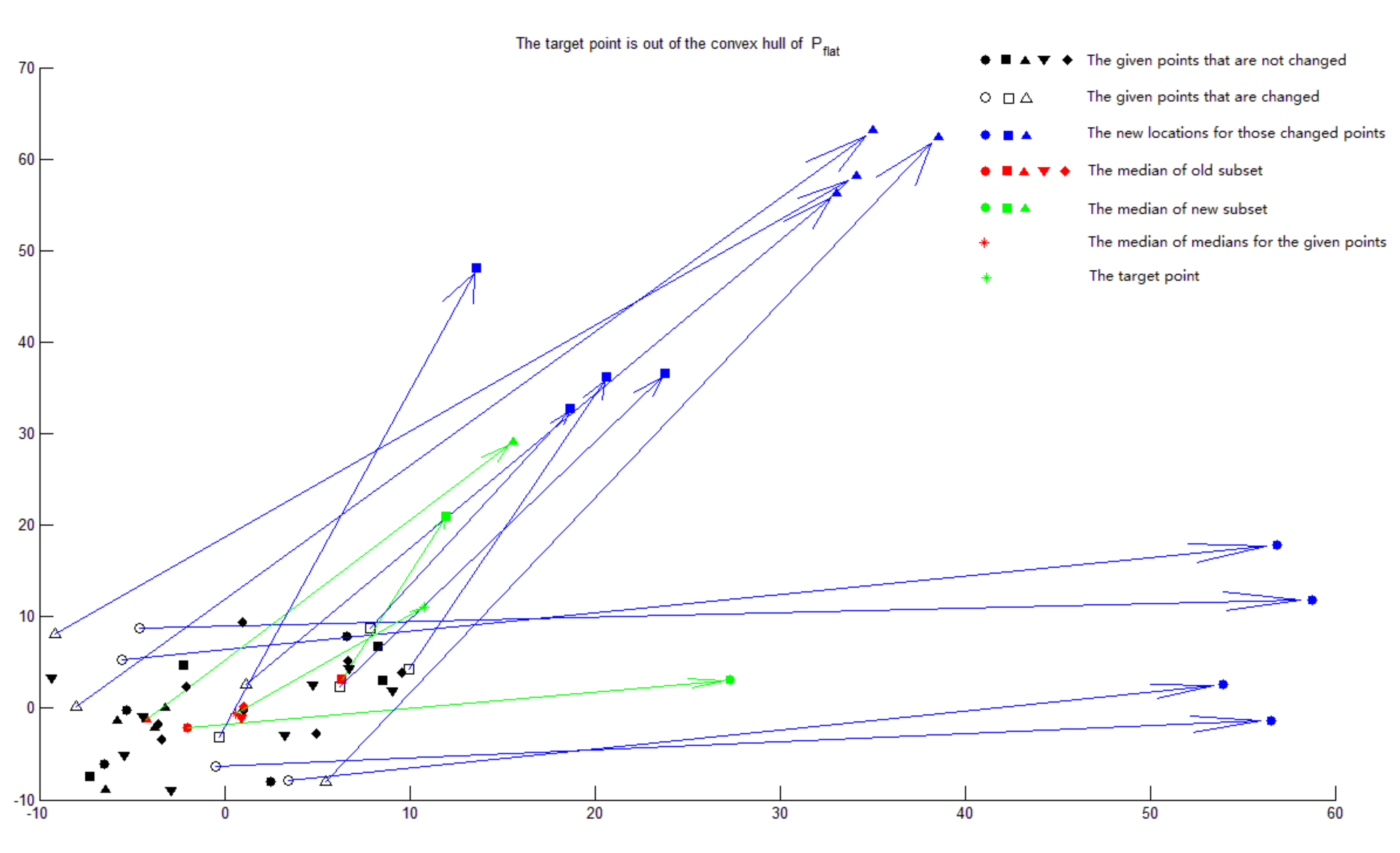}\\
  \caption{The running result for the case $n=5$, $k=8$, $(x_0,y_0)=(10.7631, 11.0663)$  in Table \ref{simulation 1}.}\label{fig2}
\end{figure*}

We illustrate the results of this process graphically for a couple of examples in Table \ref{simulation 1}; for the cases $n=5$, $k=8$, $(x_0,y_0)=( 0.9961, 1.0126)$ and $n=5$, $k=8$, $(x_0,y_0)=( 10.7631, 11.0663)$
These are shown in  Figure \ref{fig1} and  Figure \ref{fig2}, respectively.
In  these two figures, the green star is the target point. Since $n=5$, we use five different markers (circle,
square, upward-pointing triangle, downward-pointing triangle, and diamond) to represent five kinds of points. The given data $P_\all$ are shown by black points and unfilled points. Our algorithm changes those unfilled points to the blue ones, and the green points are the medians of the new subsets. The red star is the median of medians for $P_\all$, and other red points are the median of old subsets.
 So, we only changed $12$ points out of $40$, and the median of medians for the new data set is very close to the target point.

\subsection{Simulation 2 : Router Monitoring}

Suppose there are $n=100$ routers in a network, and each router monitors a stream of length $k=1000$. A router can use streaming algorithm to monitor a single percentile, for instance the frugal algorithm here~\cite{QMS2014} only needs a few bites per percentile maintained -- it does not need to monitor all.  We will consider monitoring the approximate median ($50\%$ percentile), $10\%$ percentile, and $90\%$ percentile of the stream, and sending these to a single command center.
The command center will analyze these data to determine whether an attack occurs.
In practice, command centers monitor much larger streams (values of $k$) and many more routers (values of $n$).

We use standard normal distribution to generate an array $S_i$ with $1000$ entries to simulate the $i$th stream, and assume the routers use the estimator $E_1$ to process streams, i.e. $E_1$ returns the approximate $10\%$ percentile, or $90\%$ percentile, or the median of a stream.
The command center uses the estimator $E_2$ to process the gathered data $S=(E_1(S_1),E_1(S_2),\cdots,E_1(S_n))$, and $E_2$ can return
the $10\%$ percentile, or $90\% $ percentile, or the median of $S$.
In our simulation, we compute each of these quantities exactly.
We use outliers in interval $[100,110]$ or $[-110,-100]$ to simulate attacks.

These values may represent some statistic deemed worth monitoring, say the packet length or header size after it has been appropriately normalized.

We choose $n_1$ streams, and put $k_1$ outliers from the same interval (all positive, or all negative) to each chosen stream.
Table \ref{simulation 2} shows the final output from command center for different combinations of estimators and outliers.
The first column in Table \ref{simulation 2} shows the proportion of outliers, which is equal to $\frac{n_1k_1}{nk}$.
For example, in the third row of the table, we choose 11 streams randomly and put 110 outliers drawn from [100,110] into each chosen stream,
so the proportion of outliers is $(11\times 110)/(100\times 1000)=1.21\%$.
When a value being monitored as a composite of various percentiles becomes very large (above $100$, so not from the normal distribution) we mark it \textbf{bold}.

\begin{table}
\small
\caption{The output for different combinations of estimators and outliers.} \label{simulation 2}
\centering
\begin{tabular}{|c|c|c|c|c|c|c|c|c|}
\hline
Proportion of&location of& $n_1$ & $k_1$ & $E_1$: 10\% & $E_1$: 90\% &  $E_1$: 10\% & $E_1$: 90\% & $E_1$: median \\
outliers & outliers & & & $E_2$: 10\% & $E_2$: 90\%& $E_2$: 90\% & $E_2$: 10\% & $E_2$: median\\
\hline
0\%&   &  0  &  0   &    -1.3327 & 1.3549 &   -1.2169 &    1.2254 &  -0.0085 \\
\hline
1.21\%& [100,110]&  11 & 110 & -1.3539 &\textbf{100.5666}  & -1.2033 &   1.2093  &  0.0091\\
\hline
1.21\%&  [-110,-100]& 11 & 110 &  \textbf{-100.6573}  &  1.3291  & -1.2065  &  1.2175  &  0.0021\\
\hline
10.01\%& [100,110]& 11 & 910 & -1.3364 & \textbf{108.6957} & \textbf{100.0553}  &  1.2118  &  0.0082 \\
\hline
10.01\%&  [-110,-100]& 11 & 910 &  \textbf{-108.7768}  &  1.3388 &  -1.2081  & \textbf{-100.0721}  & -0.0119 \\
\hline
26.01\%& [100,110]& 51 & 510 &    -1.3388 & \textbf{108.1641}  & -0.7794  &  1.2347 & \textbf{100.1062} \\
\hline
26.01\%&  [-110,-100]& 51 & 510 &  \textbf{-108.2083}  &  1.3163  & -1.2313  &  0.7697 &  \textbf{-100.1018} \\
\hline
46.41\%& [100,110]& 51 & 910 &   -1.3350 & \textbf{ 108.9832} & \textbf{ 100.1411}  &  1.2280 & \textbf{104.2258}\\
\hline
46.41\%&  [-110,-100]& 51 & 910 & \textbf{-109.0043}  &  1.3350  & -1.2423  & \textbf{-100.1340}  & \textbf{-104.0705} \\
\hline
\end{tabular}
\end{table}

It is shown in Table \ref{simulation 2} that for the case $E_1:10\%$, $E_2:10\%$ and $E_1:90\%$, $E_2:90\%$, we can use $1.21\%$ of outliers to change the output of $E_1$-$E_2$ estimator, since in this situation the breakdown point of $E_1$-$E_2$ estimator is $0.01$.
For the case $E_1:10\%,E_2:90\%$ and $E_1:10\%,E_2:90\%$, we can use $10.01\%$ of outliers to change the output of $E_1$-$E_2$ estimator, since in this situation the breakdown point of $E_1$-$E_2$ estimator is $0.09$.
When $E_1$ and $E_2$ both return the median of a data set, we can use $26.01\%$ of outliers to change the output of $E_1$-$E_2$ estimator, since in this situation the breakdown point of $E_1$-$E_2$ estimator is $0.25$.

This experiment illustrates how using various composite estimators with different percentiles can highlight various levels of potential distributed denial of service attacks.  For instance, if only the $E_1:10\%,E_2:10\%$ estimator is flagged, then we see a few routers have a few anomalous packets, and even though it is distributed to only about $10\%$ of routers and $10\%$ of data, we can observe it; but for the most part would be at most a warning.
If $E_1:10\%,E_2:90\%$ estimator or $E_1:50\%,E_2:50\%$ estimator is flagged, it means at least $9\%$ or $25\%$ of the packets across all routers much be anomalous, and we may see a real DDS or an early sign of one.
These are all conservative estimates.
On the other hand, if at least $10\%$ of the packets are modified on $10\%$ of routers (not too much, perhaps as little as $1\%$), then the $E_1:10\%,E_2:10\%$ estimator will definitely observe it.  And if at least $10\%$ of the packets are modified on $50\%$ of the routers (over $5\%$ of all packets), then an $E_1:10\%,E_2:50\%$ estimator will definitely observe it.
Further work is required to discover the best combination of percentiles to monitor, but using our observations about composite estimators suggests this approach which can monitor against various distributions of DDS attacks without only a few simple estimators, requiring a few bites each, at each router.

\section{Discussion}

In this paper, we define the breakdown point of the composition of two or more estimators.  These definitions are technical but necessary to understand the robustness of composite estimators; and they do not stray too far from prior formal definitions~\cite{Ham71,DG07}.
Generally, the composition of two of more estimators is less robust than
each individual estimator.  We highlight a few applications and believe many more exist.
These results already provide important insights for complex data analysis pipelines common to large-scale automated data analysis.
Moreover, these approaches provides worst case guarantees that are concrete about when outliers can or cannot create a problem, as opposed to some regularization-based approaches that just tend to work on most data.

Next we will highlight a few more insights from this work, or discuss challenges for follow-on work.


\paragraph{On the dangers of composition.}
The common case of composing two estimators, each with breakdown point of $0.5$ yields a composite estimator of $0.25$.  This means if the result is anomalous, at least $25\%$ of the data must change, down from $50\%$.  In other cases, the resulting composite estimator might yield an even smaller breakdown point of say $0.05$.  This seems like very bad news!  But for large data sets, adversarially changing $5\%$ of data is still a lot.  For instance with $1$ million data points, then $5\%$ is $50,000$, which would still be an ominously difficult task to modify.  So even a $0.05$ or $0.01$ breakdown point on large data is a useful barrier to manipulation (of the sort in our Simulation 1 below).
On the other hand, repeated composition can quickly (exponentially) decrease the breakdown point until it is dangerously low; hence we believe this new theory will play an import role in understanding the robustness and security of long data analysis pipelines.

\paragraph{Robustness and unbiasedness.}
In this paper, we focus exclusively on the robustness of estimators, but it is also important to aim for low-MSE or unbiasedness estimators.  An interesting future direction is to design estimators that are both robust (including have large onto-breakdown points) as well as other properties.  We lead this direction with a few points:
\begin{itemize} \denselist
\item Composing two unbiased estimators will typically be unbiased (some care may be needed in weighting).
\item Robustness is a worst-case analysis (protecting against adversarial data) and its claims are often orthogonal to those about low-MSE.
\item Our analysis bounds the robustness of composition of \emph{any} two (or more)  estimators.  So if other work independently shows low-MSE or low-bias properties, then we can immediately combine these works to show both.
\end{itemize}

\paragraph{Removing all subsets size $k$ constraint.}
The restriction $|P_i| = k$ (all subsets at the first level are the same size) is mainly for expositional convenience. Otherwise, there are some technical issues with reweighing points in $P_\all$ and defining the limits.
In fact, suppose $|P_i|=k_i$ for $i=1,2,\cdots,n$, $P_\all=\uplus_{i=1}^nP_i$,  $g_{E_1}(k_1) \leq g_{E_1}(k_2) \leq \cdots \leq g_{E_1}(k_n)$,
and
\begin{equation*} 
E(P_\all)=E_2\left(E_1(P_1),E_1(P_2),\cdots,E_1(P_n)\right).
\end{equation*}
Then using the
method in the proof of Theorem \ref{theorem the lower bound of the breakdown point of E1-E2}, we can obtain a result similar :
\begin{equation}
\sum_{i=1}^{g_{E_2}(n)} g_{E_1}(k_i) \leq g_E(\sum_{i=1}^{n} k_i)
\end{equation}
which is a generalization of \eqref{the lower bound of the breakdown point of E1-E2}.

\paragraph{Finite sampling breakdown point for composite estimators.}
Theorem \ref{theorem the upper bound of the breakdown point of E1-E2} provides an asymptotic breakdown point for composite estimators.   But for smaller data sets, a finite sample version is also useful and important.
Equation \eqref{the lower bound of the breakdown point of E1-E2} already gives a lower bound of the finite sample breakdown point of composite estimators.
To get an upper bound on the finite sample vesion, we can modify Theorem \ref{theorem the upper bound of the breakdown point of E1-E2}, by adding a condition $f_{E_1}(k)=g_{E_1}(k)+C$ where $C$ is a positive constant.  Then there is also an annoying off-by-one error on $g_{E_2}$ (see eq \eqref{frac{g_{E_1}(k)}{k} frac{g_{E_2}(n)}{n}leq frac{g_E(nk)}{nk}leq frac{f_{E_1}(k)}{k} frac{(g_{E_2}(n)+1)}{n}}), so the result would be something like
\[
g_{E_1}(k)g_{E_2}(n) \leq g_E(nk) \leq (g_{E_1}(k) +C)(g_{E_2}(n) + 1),
\]
and it is not completely tight.  We leave providing a tight bound (up to these constants) as an open question.

\newpage
\small
\bibliographystyle{abbrv}
\bibliography{estimator_full_version}
\normalsize

\newpage
 \appendix



\end{document}